\documentclass[conference]{IEEEtran}
\IEEEoverridecommandlockouts
\usepackage{cite}
\usepackage{amsmath,amssymb,amsfonts}
\usepackage{algorithmic}
\usepackage{graphicx}
\usepackage{textcomp}
\usepackage{arabtex}
\usepackage{utf8}
\setcode{utf8}
\usepackage[utf8x]{inputenc} 
\usepackage[T1]{fontenc}    % use 8-bit T1 fonts
\usepackage{hyperref}       % hyperlinks
\usepackage{url}            % simple URL typesetting
\usepackage{booktabs}       % professional-quality tables
\usepackage{amsfonts}       % blackboard math symbols
\usepackage{nicefrac}       % compact symbols for 1/2, etc.
\usepackage{microtype}      % microtypography
\usepackage{lipsum}		% Can be removed after putting your text content
\usepackage{graphicx}
\usepackage{doi}
\usepackage{float}
\usepackage{subfig}

\usepackage{multirow}
\usepackage{booktabs, multirow} % for borders and merged ranges
\usepackage{soul}% for underlines
\usepackage[table]{xcolor} % for cell colors
\usepackage{changepage,threeparttable} % for wide tables
\usepackage{adjustbox}

\graphicspath{ {./} }

\def\BibTeX{{\rm B\kern-.05em{\sc i\kern-.025em b}\kern-.08em
    T\kern-.1667em\lower.7ex\hbox{E}\kern-.125emX}}
\begin{document}

\title{A Multi-Purpose Audio-Visual Corpus for Multi-Modal Persian Speech Recognition: the Arman-AV Dataset}

\author{
\IEEEauthorblockN{Javad Peymanfard\,\IEEEauthorrefmark{1},
Samin Heydarian\,\IEEEauthorrefmark{1}\textsuperscript{\textsection},
Ali Lashini\,\IEEEauthorrefmark{1}\textsuperscript{\textsection}, 
Hossein Zeinali\,\IEEEauthorrefmark{2},\\
Mohammad Reza Mohammadi\,\IEEEauthorrefmark{1}, and
Nasser Mozayani\,\IEEEauthorrefmark{1}}
\vspace{2mm}
\IEEEauthorblockA{\IEEEauthorrefmark{1}\,School of Computer Engineering\\
Iran University of Science and Technology, Tehran, Iran \\
Email: \texttt{javad\_peymanfard@comp.iust.ac.ir, samin\_heydarian@comp.iust.ac.ir}\\
\texttt{a\_lashini@comp.iust.ac.ir, mrmohammadi@iust.ac.ir }\\
\texttt{mozayani@iust.ac.ir}
}
\vspace{2mm}
\IEEEauthorblockA{\IEEEauthorrefmark{2}\,Department of Computer Engineering \\
Amirkabir University of Technology, Tehran, Iran \\
Email: \texttt{hzeinali@aut.ac.ir}
}
}

\maketitle

\begingroup\renewcommand\thefootnote{\textsection}
\footnotetext{These authors contributed equally.}
\endgroup

\begin{abstract}
In recent years, significant progress has been made in automatic lip reading. But these methods require large-scale datasets that do not exist for many low-resource languages. In this paper, we have presented a new multipurpose audio-visual dataset for Persian. This dataset consists of almost 220 hours of videos with 1760 corresponding speakers. In addition to lip reading, the dataset is suitable for automatic speech recognition, audio-visual speech recognition, and speaker recognition. Also, it is the first large-scale lip reading dataset in Persian. A baseline method was provided for each mentioned task. In addition, we have proposed a technique to detect visemes (a visual equivalent of a phoneme) in Persian. The visemes obtained by this method increase the accuracy of the lip reading task by 7\% relatively compared to the previously proposed visemes, which can be applied to other languages as well.
\end{abstract}


\section{Introduction}

Automatic speech recognition (ASR) is a task to understand speech from audio signals. This task has been developed over the years and got mature enough to be used on any device. But the trained models, apart from the high ability of speech recognition, have weaknesses in special conditions like environments with loud noises. That's where audio-visual speech recognition (AVSR) comes in to overcome this limitation. AVSR uses visual information alongside audio in order to decode speech more effectively in noisy environments like inside cars. AVSR is more like human comprehension, which uses visual and audio perception to understand each other. Lip reading is another task which is only consuming visual information.

The traditional approaches~\cite{matthews2002extraction, MORADE20145181, zhou2011towards} used a two-stage algorithm to deal with such problems. In the first stage, a hand-crafted feature extractor, extract useful features from lip movements alongside another feature extractor that extracts feature from audio signals and then fuse them. In the second stage, a classifier such as the hidden Markov model or artificial neural networks is used to classify digits, characters, etc. However, in the last few years, the availability of large public datasets and emerge of deep neural networks had a substantial impact on this field. Deep learning approaches usually consist of two parts, which are front-end and back-end, like traditional ones except that here they are end-to-end trainable. For front-end part usually use convolution neural networks to extract visual and audio features and temporal networks like RNNs, attention model, and transformers on the back-end side to model temporal information. Recently, methods such as~\cite{zhao2020hearing, afouras2020asr, ren2021learning} used knowledge distillation to train lip reading and AVSR models. To do this, usually use the ASR model as the teacher and the lip reading model as the student.

The primary dataset was collected under laboratory conditions~\cite {cooke2006audio, anina2015ouluvs2} - Usually, a person stands in front of the camera and read some words or sentences in a quiet place at normal speed. Emerge of deep learning and automatic pipeline led to building datasets in the "wild" condition~\cite{chung2017lip, afouras2018lrs3, shillingford19_interspeech} which is larger and also more challenging and close to the real condition than before. Unlike the old datasets which were restricted to digit, character, or phrase classification, nowadays, datasets are collected for word-level classification and sentence-level AVSR. Unfortunately, most of these datasets are in English and there are a few datasets witches come in other languages~\cite{zhao2019cascade}. In this paper, we will more concentrate on sentence-level AVSR, but we did not limit our dataset to just speech recognition tasks. We make our dataset in a way that can be used for different tasks. In the following sections, we will explain more about it.

Currently, LRS2~\cite{son2017lip} is the most wildly used audio-visual dataset. The dataset contains more than 240 hours of videos and about 118,000 utterances. Also, they used BBC news and talk shows as sources for their dataset. The dataset is publicly available to the research community. LSVSR~\cite{shillingford19_interspeech} is the largest dataset with over 3800 hours of data collected from YouTube-uploaded videos. This dataset was collected by Google DeepMind and, unfortunately, they did not make it publicly available. 

In this paper we propose a novel large-scale dataset that is collected in the "wild" condition from reviews, movies, etc. on the Aparat website. The dataset contains over 220 hours of videos from 1760 Persian celebrities. Also, we provide the name of the celebrity in each sample as labels to be used for different purposes, such as speaker recognition. To the best of our knowledge, this is the largest audio-visual dataset in the Persian language. 

Since this is an audio-visual dataset, it could be used for a number of different applications such as automatic speech recognition, lip reading, speaker recognition, audio-visual speech synthesis, etc. 

The organization of this paper is as follows. In section 2, we look at some of the most recent approaches and datasets. In section 3, we discuss the statistic of the data and the pipeline that we build for collecting data. In section 4, we will use a base model to evaluate our dataset. Finally, conclude the paper in section 5.

\section{Related Works}

\begin{table*}[ht]
    \begin{adjustbox}{width=1\textwidth}
    \begin{threeparttable}[b]
    \begin{center}
    \caption{Statistical comparison between well-known datasets.}
    \label{tab:datasets_statistic}
    \begin{tabular}{|c|c|ccccccc|cc|ccccc|c|}
    \midrule
      \textbf{Category} & \textbf{Dataset} & \multicolumn{7}{c|}{\textbf{General info}} & \multicolumn{2}{c|}{\textbf{Transcription info	}}& \multicolumn{5}{c|}{\textbf{Video info}}&
      \textbf{Speaker info}\\
      \hline
      Category	& Name & Year &	Lang. & Task &	Classes	& Utter. &	Availability &	Source & Vocab.	& Words & Num.	& Duration	& Resolution & FPS &	View(°) & Num.\\
      \midrule
      
      & SFAVD~\cite{naraghi2013sfavd} & 2013 & FA & Sent. & - & 600 & - & Lab environment & ~1000  & - & - & - & 131×105 &	30	& Frontal & 1\\
      
      & LRW~\cite{chung2017lip}	& 2016	& ENG & Words & 500 & 400K & Avail.*\tnote{1}& BBC Programmes& 500 & 400K & - & - & 256×256	& 25 & -30~30 & 1000+\\
      
      & LRS2~\cite{son2017lip} & 2017 & ENG & Sent. & - & 118K & Avail.*& BBC Programmes& 174K & 807K & - & 246h & 160×160 & 25 & -30~30 & 1000+\\
      
      & MV-LRS~\cite{chung2017lipmv} & 2017 & ENG & Sent. & - & 74K & Avail.*& Programs& 14K & - & - & ~165h & 160×160 & 25 & -90~90 & 1000+\\
      
       \multirow{1}{*}{Audio-Visual} & VLRF~\cite{fernandez2017towards} & 2017 & SPA & Sent. & - & 600 & Public & Lab environment & 1374 & 10K & - & 180min & 1280×720 & 50 & Frontal & 24\\
       
      \multirow{1}{*}{ Speech } & LRS3-TED~\cite{afouras2018lrs3} & 2018 & ENG & Sent. & - & 165K & Avail.*&TED Talks& ~57K & - & - & ~475h & 224×224 & 25 & -90~90 & 1000+\\
      
      \multirow{1}{*}{Recognition} & LSVSR~\cite{shillingford19_interspeech} & 2018 & ENG & Sent. & - & 2,934K & Private & YouTube& 127K & - & - & 3,886h & 128×128 & 30 & -30~30 & 1000+\\
      
       & LRW-1000~\cite{yang2019lrw} & 2019 & CHI & Words & 1000 & 718K & Avail.* & Broadcast news & 1K & - & - & - & Distributed & 25 & -90~90 & 2000+\\
       
      & CMLR~\cite{zhao2019cascade} & 2019 & CHI & Words & - & 102K & Public & News Broadcast & - &- &  - & - & - & - & - & 11\\
      & LRWR~\cite{egorov2021lrwr} & 2021 & RUS & Words & 235 & - & - & YouTube & 235 & 117K & - & - & 112×112 & 25 & 0~20 & 135\\
      & GLips~\cite{schwiebert2022multimodal} & 2022 & GER & Words & 500 & - & Public & Hessian Parliament & 500 & 250K & 250K & - & 256×256 & 25 & - & ~100\\
      & RUSAVIC~\cite{ivanko2022rusavic} & 2022 & RUS & - & 62 & 62 & Avail.*& Vehicle environment & - & - & 200 & - & 1920×1080 & 60 & -30~30 & 20\\
      & PLRW~\cite{peymanfard2022word} & 2022 & FA & Words & 500 & 244K & Public & Aparat & 500 & - & 244K & 30h & 224×224 & 25 & - & 1800\\
      & Our Dataset & 2022 & FA & Sent. & - & 89K & Public & Aparat & 42K & 2.5M & 89K & 220h & 224×224 & 25 & - & 1760\\
      \midrule
      \multirow{1}{*}{Acitive Speaker}  
      & AVA~\cite{roth2020ava} & 2019 & - & - & 3 & - & Public & YouTube &	- & - &	188 & 38.5h	& 128×128 &	- & - & -\\
      \multirow{1}{*}{Detection}  
      & ASW~\cite{kim2021look}	& 2021 & - & - & 2 & - & - & VoxConverse dataset & - & - & 212 & 30.9h & - & - & - & -\\
      \midrule
       
      & Voxceleb~\cite{Nagrani17} & 2017 & ENG & - & 1251 & 153K & Public & YouTube &	- & - &	22K & 352h &	- & - &	- & 1,251 \\
      \multirow{1}{*}{Speaker} 
      & Voxceleb2~\cite{Chung18b} & 2018 & ENG & - & 6112 &	1,128K &	Public & YouTube & - & - & 150K & 2442h & - & - & - & 6,112\\
      \multirow{1}{*}{Recognition} 
      & DeepMine~\cite{zeinali2018deepmine} & 2019	& FA\&ENG &	- &	1850 & 544K	& Avail.*& Crowdsourcing	& - & - & 540 &	480h & - & - & - &1,850\\
      & CN-Celeb~\cite{fan2020cn} & 2019 & CHI & - & 1000 & 130K & Public & bilibili.com & - & - & - & 273.73h & - & - & - & 1,000\\
\midrule
    \end{tabular}

     \begin{tablenotes}
       \item [1] Available by contact.
     \end{tablenotes}
    \end{center}
    \end{threeparttable}
    \end{adjustbox}
\end{table*}

In this section, we will review the two tasks of speech recognition and speaker recognition. Since we have introduced a new dataset and tested the performance of the baseline models on it, we will explore the related works from the perspective of the proposed methods and datasets.

\subsection{Methods}
\subsubsection{Speech Recognition}
\textbf{Lip reading.}
There is a large body of research on automated lip reading. The advent of deep learning and the availability of large-scale datasets cause massive progress in the automation of this field. Here we discuss the works which focused on the sentence-level task in lip reading as an open set character. For sentence-level recognition, the recent works can be divided into approaches.
The first approach utilizes Connectionist Temporal Classification (CTC). The model predicts frame-wise characters and tries to maximize and marginalizes all possible paths to find optimal alignment between prediction and ground truth. An example based on this approach is LipNet~\cite{DBLP:journals/corr/AssaelSWF16}. LipNet was the first sentence-level lip reading model that used spatiotemporal CNNs as the front-end, followed by Bidirectional Gated Recurrent Unit (Bi-GRU) as the back-end of architecture, and employed CTC loss to train the network. Shillingford \textit{et al.}~\cite{shillingford19_interspeech} proposed the V2P model closest to previous work. The model outputs a sequence of phoneme distributions and uses CTC loss at train time. At inference time, a decoder based on finite-state transducers (FSTs) maps a sequence of phoneme distributions to a word sequence.
 
The second approach follows the sequence-to-sequence model, in which output characters are conditioned based on each other, unlike CTC-based models. WAS~\cite{chung2017lip} was the first sequence-to-sequence model. The model consists of two key components called the image encoder Watch and the character decoder Spell and has a unique dual attention mechanism. Chung \textit{et al.}~\cite{chung2017lipmv} extended the WAS model to the MV-WAS model that can decode visual sequences across all poses and show that it is possible to read lips in profile, but the standard is inferior to reading frontal faces.  Peymanfard \textit{et al.}~\cite{peymanfard2022lip} suggested external viseme decoding, which divides the sequence-to-sequence model into two stages, video to viseme and viseme to character, respectively. The network outputs character sequence given viseme through external text data. Afouras \textit{et al.}~\cite{afouras2021sub} proposed a visual transformer pooling (VTP) module that learns to track and aggregate the lip movements representation. This work also utilizes a wordPiece tokenizer (one of the subword algorithms) to learn the language easily, resulting in time and memory efficiency
 
Petridis \textit{et al.}~\cite{petridis2018audio} presented a hybrid architecture and used a joint decoder including RNN, attention, and CTC. Afouras \textit{et al.}~\cite{afouras2018deep} took a hybrid approach as well but suggested replacing RNN with a transformer. This work proposed a sequence-to-sequence and CTC on top of the transformer self-attention as the back-end. Both models use spatiotemporal CNNs, followed by ResNet as a common front-end. The extension of~\cite{petridis2018audio} is~\cite{ma2021end} that employed the conformer variant of the transformer to model both local and global dependencies of an image or audio sequence.

There is a trend in lip reading that benefits from the knowledge distillation area to improve performance by extracting information from the audio-only counterpart of datasets. In this approach, the Automatic Speech Recognition (ASR) model plays a role as the teacher for the Visual Speech Recognition (VSR) model as the student. The examples of this trend are~\cite{zhao2020hearing},~\cite{afouras2020asr} and~\cite{ren2021learning}.\\
\textbf{Automatic Speech Recognition.} In ASR, self-supervision plays the main role in the state-of-the-art models. This approach has two steps, pre-training and fine-tuning.
In the pre-training step, the model learns powerful speech representation from large amounts of unlabeled data through a pretext task. Then, the output of the previous step (speech representation) is fed to an acoustic model to be fine-tuned for a downstream task. In wav2vec~\cite{schneider2019wav2vec}, the pre-training step has two networks named encoder and context. The encoder network converts raw audio input to feature representations, and the context network converts feature representations to contextualized features. Both networks are convolutional networks. The objective of this model is defined by contrastive loss between future steps and distractors. The vq-wav2vec~\cite{baevski2019vq} is the improved version of the previous work, which benefited from the BERT model. This work is also composed of two stages. In the first stage, the continuous audio feature
representations are discretized via Gumbel-Softmax or k-means clustering quantization methods. Thanks to this, the speech data takes a structure similar to language, and the BERT model can be applied. So in the second stage, the discretized representations are fed to a BERT model for contextualized representations. The wav2vec 2.0~\cite{baevski2020wav2vec} showed better results by joint learning of the two-stages pre-training pipeline mentioned above. The model is pre-trained by new contrastive loss between contextualized output and quantized representation. The HuBERT~\cite{hsu2021hubert} is very similar to wav2vec 2.0 in model architecture but different in the training process. First, HuBERT uses the cross-entropy loss for model optimization. Second, Data discretization is done through a k-means algorithm instead of a quantization process. Third, The training process alternates between the hidden units discovery and target prediction. The model re-uses embeddings from the BERT encoder in the clustering step.\\
\textbf{Audio-Visual Speech Recognition.}
The task of audio-visual speech recognition is lip reading with the presence of audio. Shi \textit{et al.}~\cite{shi2022learning} presented the audio-visual counterpart of HuBERT model as an AV-HuBERT. Also, this work reported the result of visual HuBERT performance for the lip reading task. The works like~\cite{shillingford19_interspeech},~\cite{petridis2018audio} and~\cite{afouras2018deep} addressed both the VSR and the AVSR problems, which are later explained in the lip reading section.
\subsubsection{Speaker Recognition}
Since this is a classification problem, the general models consist of two modules. The first is a feature extractor module and the second is a classification module. Depending on the input data, the feature extractor could be a convolutional network like VGG-M or ResNet~\cite{Chung18b} or any other neural network that could be used for sequential data. For classification, a fully connected network is the leading choice, and the output size is the number of speakers in the dataset. Since the input data is an audio file, input data to the network could be a spectrogram~\cite{Chung18b} of data or the output of a feature extraction algorithm like MFCC, etc., which usually is a 1D feature vector of data.

\subsection{Datasets}

\begin{figure*}[t]
  \centering
  \includegraphics[width=0.95\textwidth]{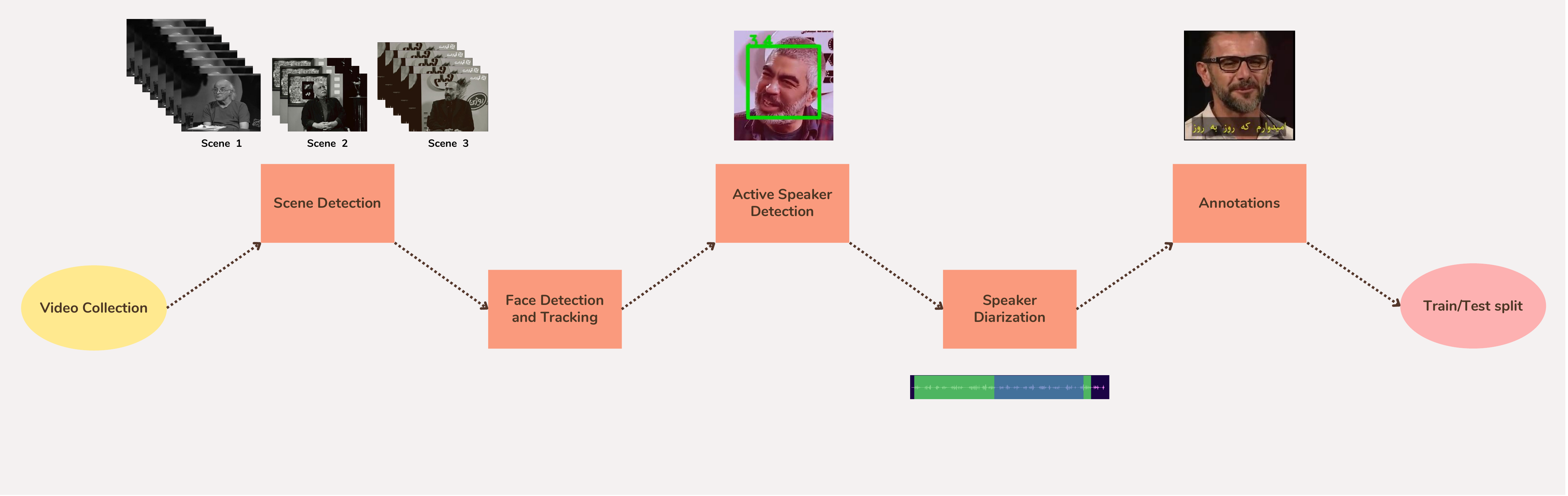}
  \caption{Data collection pipeline.}
  \label{fig:data_collection_pipeline}
\end{figure*}

\begin{figure*}[t]
  \centering
  \includegraphics[width=0.9\textwidth]{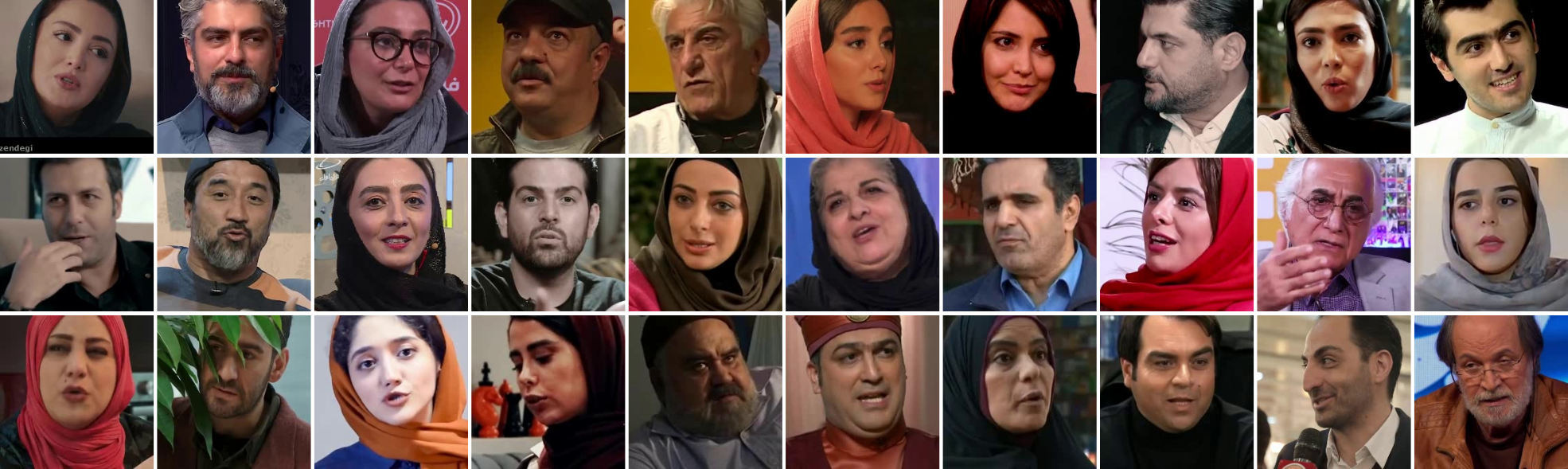}
  \caption{Our dataset samples.}
  \label{fig:dataset_samples}
\end{figure*}

\subsubsection{Speech Recognition}
As we know, the progress in deep learning is due to large datasets. The performance of lip reading systems is affected by position, illumination and speaker diversity, so the quality and quantity of the dataset are important. In recent years some datasets like LRS2~\cite{son2017lip}, LRS3-TED~\cite{afouras2018lrs3} and LSVSR~\cite{shillingford19_interspeech} have tried to increase the number of utterances and video hours and  MV-LRS~\cite{chung2017lipmv} has tried to cover more face angles. Table \ref{tab:datasets_statistic} shows three categories of datasets named Audio-Visual Speech Recognition, Active Speaker Detection and Speaker recognition. The table has general, transcription, video and speaker information columns for each dataset. We can see useful statistics in this information, such as the type of task run on the dataset, the source of the collected videos and the number of speakers in the videos. The language of most of the datasets in this table is English, but fortunately, other languages have been published in recent years, like CMLR~\cite{zhao2019cascade}, LRWR~\cite{egorov2021lrwr} and GLips~\cite{schwiebert2022multimodal}. Since 2013, no Persian dataset has been published for lip reading until now; we released Audio-Visual Speech Recognition in Persian (Arman\_AV) with 89k utterances, 220 hours and 1760 speakers.

\subsubsection{Speaker Recognition}
Most of the datasets in this field are collected under laboratory conditions and are not available to be used freely for research purposes. One of the first free datasets which were collected under the "wild" conditions is the Speakers in the Wild (SITW)~\cite{McLaren2016TheSI} dataset. This dataset contains multimedia content of 299 speakers with hand-annotated speech samples. VoxCeleb~\cite{Nagrani17} is a large-scale dataset that contains speech samples of over 1,000 celebrities with more than 100,000 utterances, extracted from YouTube videos; this dataset is gender-balanced (45\% of the speakers are female), along with speakers with various accents, racial backgrounds, and ages. The dataset's creators included information about each speaker's gender and country of origin from Wikipedia. Videos contained in the dataset are recorded in a huge number of challenging environments such as quiet and noisy studio interviews, open-air stadiums, etc., and all are debased with real-world commotion, consisting of giggling, covering discourse, foundation chatter, etc. Also, another goal of the authors of the dataset was to propose a pipeline to create fully automated datasets. VoxCeleb 2~\cite{Chung18b} which came shortly after VoxCeleb (and is very similar but much larger) is the largest dataset in this field which includes more than six thousand speakers that speak more than a million utterances. The three mentioned datasets are in English. CN-Celeb~\cite{fan2020cn} is another large-scale dataset which contains over 130,000 utterances and is very similar to the VoxCeleb dataset except in the three following aspects: first, CN-Celeb covers 11 genres of speech such as singing, entertainment, etc., which is more than VoxCeleb. Second, CN-Celeb concentrated on Chinese celebrities, containing videos of 1,000 celebrities. And last but not least is that the dataset is not automated completely, and they considered human supervision for the dataset. To the best of our knowledge, there is no suitable Persian dataset in this field.

\section{Dataset}

\subsection{Automated Pipeline}
In this subsection, we see the steps of data collection. Figure \ref{fig:data_collection_pipeline} shows the overview of these steps.\\
\textbf{Step 1: Video Collection.}
    We pick three kinds of videos for our dataset described in the following segments.
    \begin{enumerate}
      \item \textbf{Interviews.}
        Interviews and biographies are well-suited for audio-visual datasets. The main goal of these videos is to speak to artists, politicians or famous people in general. These kinds of videos are divided into two groups. In the first group, the shows include a host and narrator (which is not desirable in this case), and the second group consists of those that don't have them.
        % So with videos in the first group, the host should be eliminated.  Since our purpose is an audio-visual dataset, the narrator should be eliminated too. In the second group, some breaks should be recognized and removed.
      \item \textbf{Series and Movies.}
      These kinds of videos can be used if suitable preprocessing is applied to them. The main challenges are dealing with different directions and angles of the camera (When a speaker is talking or is not in the shot all of the time, unlike in the interviews). Also, the entire video has a significant amount of silence or music. Another challenge is multi-speaker simultaneous speech. In general, this type of video is not ideal for collecting data because the percentage of redundant data in them is relatively high. In the investigation we did, less than 10\% of the input data is considered appropriate.
      \item \textbf{Vods.}
      One of the other ideas to collect more data is different keyword searching in ugcs. Then we review the resulting videos and choose the top keywords for the dataset.
    \end{enumerate}

    To obtain data, multiple search terms are used

\textbf{Step 2: Scene Detection.}
      In this stage, we want to detect each scene. For this purpose, the PySceneDetect\footnote{\url{https://github.com/Breakthrough/PySceneDetect}} is utilized. In this algorithm, we subtract the pixel values of two consecutive frames to detect different scenes. This difference value is calculated in HSV colour space.
      % If the difference is more than the specific threshold then the new scene is discovered. The threshold is obtained empirically (heuristically).
    
\textbf{Step 3: Face Detection and Face Tracking.}
      Now each video is divided into smaller pieces called a scene. Up to this point, the video had a temporal clipping. Note that we need frames in which we see faces. So the S\textsuperscript{3}FD model~\cite{zhang2017s3fd} and algorithm based on IoU~\cite{tao2021someone} are used for face detection and face tracking respectively. Depending on the number of faces found in each part of the video, each video may be split into one or more videos or removed if there is no face.

\textbf{Step 4: Active Speaker Detection.}

    In this step, we automatically found the parts of the video that contain the speaker's face.
    
\textbf{Step 5: Speaker Diarization.}
    As stated, up to stage 4, the video parts including a speaker with a constant shooting angle have been chosen. But there is another potential challenge
    % if the non-active speaker and the active speaker (To whom the picture belongs) speak simultaneously. It happens more
    in interviews, a situation where the host and the guest talk at the same time. To address this problem, we can use Speaker Diarization\footnote{\url{https://github.com/taylorlu/Speaker-Diarization}} based on UIS-RNN.
    % This algorithm clusters the input audio, and each cluster has the speaker's ID.
    
\textbf{Step 6: Annotations.}
    As we mentioned before, the source of our dataset videos is Aparat. The videos on this video hosting website lack subtitles, so we utilize the commercial Aipaa's\footnote{\url{https://aipaa.ir/}} ASR service to construct rough sentence-level transcripts of the videos.
    % It is worth mentioning that the accuracy of transcriptions in this dataset is phoneme-level (The smallest unit of speech that can differentiate one word from another).

\textbf{Step 7: Face Recognition and Dataset Split.}
    We use ArcFace~\cite{deng2019arcface} and produce face feature embeddings for face recognition.
    % Then the clustering algorithm is applied to these face feature embeddings, and now we have a specific cluster for each identity.
    This stage aims to create a dataset with a speaker-independent property. A dataset like this is beneficial for applications like lip reading. 
    % Since we have a list of identities, our policy to make a test set is to pick speakers who are not in the train set.
    It should be noted that the results of the face recognition algorithm were checked manually, and its errors were corrected.
    
\subsection{statistics}
First of all, the dataset will be available as mp4 face-cropped videos with 224*224 pixels resolution and a frame rate of 25 FPS. We leveraged Aparat (a Persian video-sharing website like YouTube) as a source for our dataset. The collected videos came from different channels and programs, which means our dataset contains various kinds of challenging environments and circumstances. The outcome of our work comprises 220 hours of video, which contains over 89 thousand utterances and 2.5 million words of 1760 celebrities. We split our dataset into train/ validation, and test sections which each contain  211, 9 hours of data. Figure \ref{fig:dataset_samples} shows some examples of the video samples. 
% As Figure \ref{fig:gender_distribution} shows, the distribution of males and females is not in balance, about three-quarters of the speakers are male. 
In table \ref{tab:datasets_statistic}, in addition to the statistics of our dataset, we have also presented the statistics of other recent datasets. 

\section{Viseme Analysis}

The viseme is a significant challenge in the lip reading problem. Visemes consist of a set of phonemes spoken by the same form of lips and aren’t exactly known for any languages. This mapping may even be different in various speakers. However, some research has addressed the issue in languages such as English and Persian. Some proposed models for said languages. In this study, we proposed a method to automatically identify the Persian visemes on large-scale data that one can apply to other languages by only having the appropriate dataset of the interested language.
Some works have addressed the visemes for the Persian language ~\cite{aghaahmadi2013clustering}. Collecting a set of laboratory datasets and considering the similarities in speaking the various phonemes, the authors in~\cite{aghaahmadi2013clustering} proposed a categorization method for phonemes to identify the visemes in the language. Combining the deep learning technique to automatically extract the features and clustering in this study, we introduced a method to categorize the phonemes. Then, we compared a viseme-based model for lip reading with these visemes and the visemes addressed in previous works.
Employing the Kaldi tool, we first determined the phoneme level transcription for each sample in this experiment. Each phoneme spoken by the speaker is known every 30 milliseconds in this transcription. Then, we employed the av-Hubert pre-trained model to obtain the visual embedding for each phoneme, which was used as a feature for our clustering using the k-means algorithm. In this step, we expected the phonemes having the same visual features, like the same lip movements and forms, to be grouped in the same category.
We used the method in~\cite{peymanfard2022lip} to train our model. Employing the two models of the video-to-viseme and the viseme-to-character models and finally merging the two, we obtained the lip reading model. One benefit of such a method is that one can use the textual data to improve the quality of the lip reading model. However, the model requires a proper grapheme to phoneme (G2P) model to convert the Persian text into the phonemes and then convert it into the visemes, for which we used an appropriate G2P. For the video-to-viseme model dataset, we used the phoneme level transcription to convert the phonemes into visemes with the help of its corresponding mapping. We also employed about 1 million utterances for the visemes into character conversion datasets. In this regard, we first employed a G2P model to obtain the phoneme sequence for each utterance and then used the phoneme-to-viseme mapping to extract the viseme sequence for each one. To train the viseme-to-character model, we used the viseme sequence as the input, whereas the persian text was considered the desired output.
Employing the attention mechanism to implement the video-to-viseme model, we used a seq2seq one having a two-layer GRU. Furthermore, we employed the same architecture with fewer parameters and a different input type to train our viseme-to-character model.
As seen in table \ref{tab:results_vsr}, we obtained a higher precision in the lip reading problem by employing the new proposed model to identify the Persian visemes for which we believe this phoneme-to-viseme mapping to be more appropriate. Furthermore, one can use the method to identify visemes in any given language.
All the data and their related transcriptions are publicly available.

\begin{table*}[ht]
 \caption{Lip reading results on our dataset }
%  \vspace{2mm}
  \centering
  \begin{adjustbox}{width=1\textwidth}
  \begin{tabular}{c | c |c |c|c}
    \toprule
    \toprule
    Viseme Mapping     & CER (Greedy decoding)  & WER (Greedy decoding) & CER (Beam search decoding)  & WER (Beam search decoding) \\
    \midrule
     Traditional~\cite{aghaahmadi2013clustering} & \%54.04 & \textbf{\%76.32}  & \%52.19 & \%73.24  \\
     AV-HuBERT + K-means (Ours) & \textbf{\%51.24} & \%76.71 & \textbf{\%48.54} & \textbf{\%72.83}   \\
    \bottomrule
    \bottomrule
  \end{tabular}
  \label{tab:results_vsr}
  \end{adjustbox}
\end{table*}

\begin{figure}%
    \centering
    \subfloat[\centering ]{{\includegraphics[width=8cm]{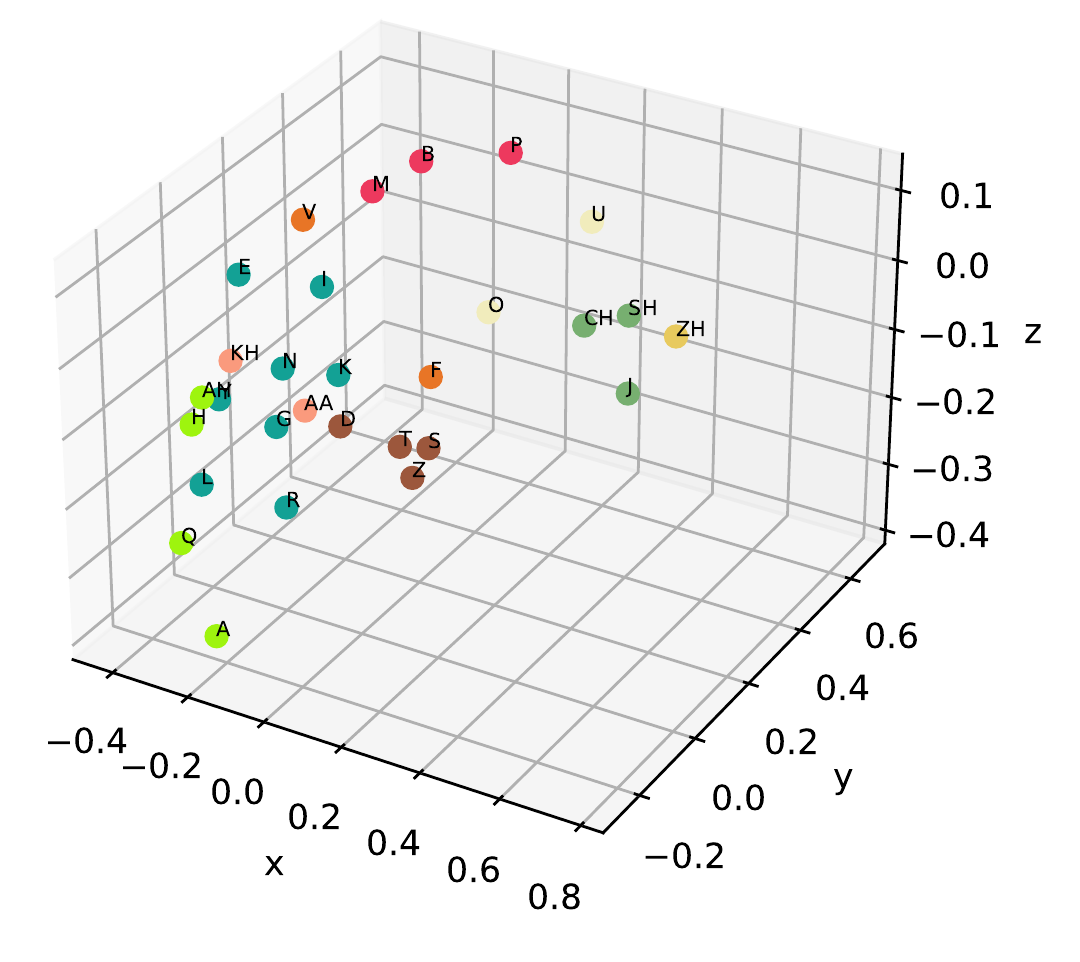} }}
    \qquad
    \subfloat[\centering ]{{\includegraphics[width=8cm]{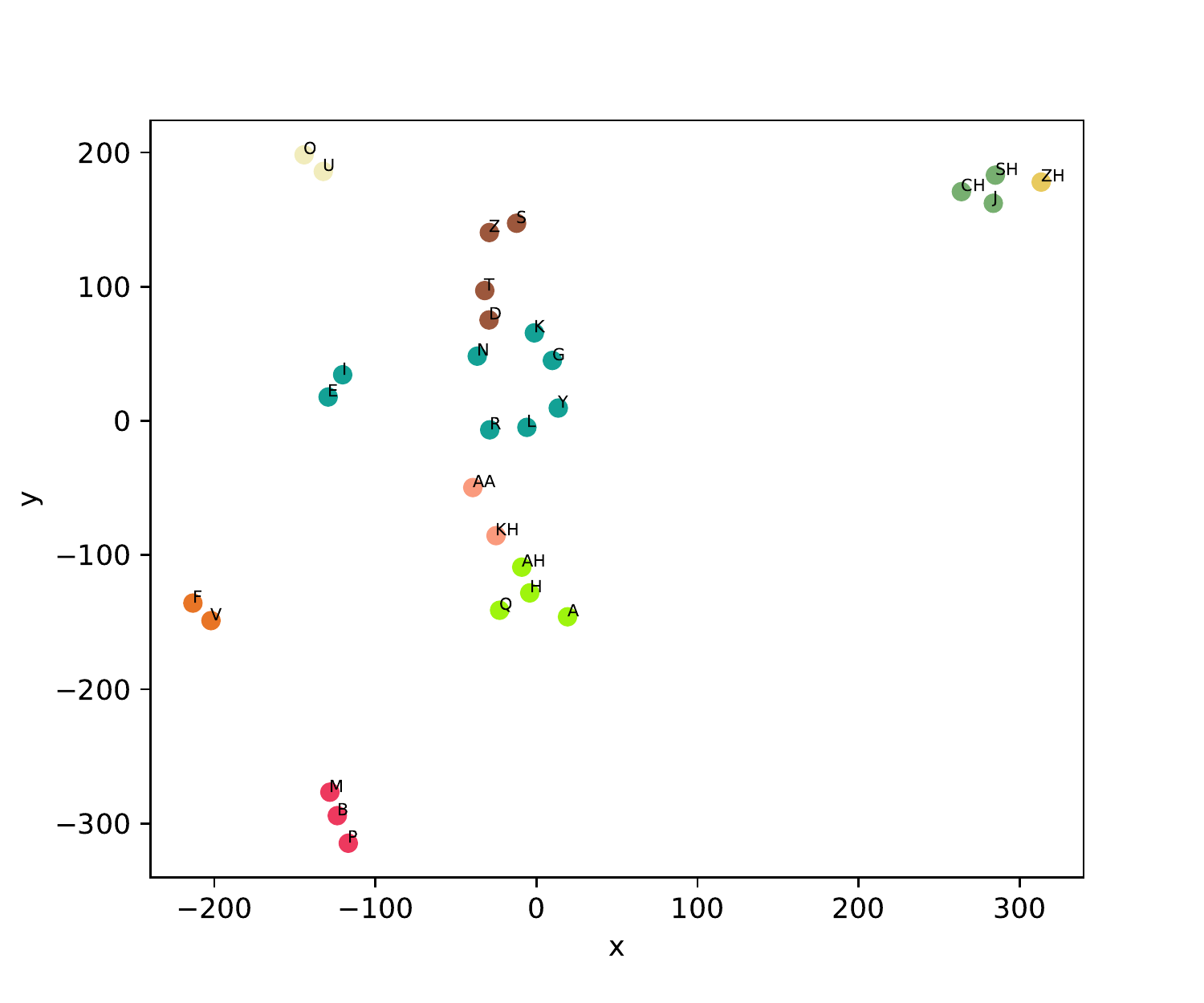} }}
    \caption{2D and 3D projection of AV-HuBERT for persian phonemes using our dataset. a) 3D projection of embedding of phonemes using PCA. b) 2D projection of embedding of phonemes using tSNE.}
    \label{fig:2&3dprojection}
\end{figure}

% \begin{figure}[t]
%   \centering
%   \includegraphics[width=0.5\textwidth]{embedding_pca_3c.pdf}
%   \vspace{-6mm}
%   \caption{No caption}
%   \label{fig:embedding_pca_3c}
% \end{figure}

% \begin{figure}[t]
%   \centering
%   \includegraphics[width=0.5\textwidth]{embedding_TSNE_2d_pp3.pdf}
%   \vspace{-6mm}
%   \caption{No caption}
%   \label{fig:embedding_TSNE_2d_pp3}
% \end{figure}

\begin{table}
 \caption{Persian visemes using clustering and AV-HuBERT}
  \centering
  \begin{tabular}{lc}
    \toprule
    cluster id     & phonemes   \\
    \midrule
    \midrule
    1 & /F/ /V/    \\
    \midrule
    2 & /AA/ /KH/  \\
    \midrule
    3 & /B/ /P/ /M/  \\
    \midrule
    4 & /ZH/ \\
    \midrule
    5 & /CH/ /JH/ /SH/  \\
    \midrule
    6 & /O/ /U/ \\
    \midrule
    7 & /S/ /Z/ /D/ /T/ \\
    \midrule
    8 & /A/ /'/ /H/ /GH/  \\
    \midrule
    9 & /G/ /K/ /E/ /L/ /I/ /Y/ /N/ /R/ \\
    \midrule
    10 & /sil/ \\
    
    \midrule
    \bottomrule
  \end{tabular}
  \label{tab:ourvisemes}
\end{table}

\section{Experiments}
Here, the baseline methods are explained. We demonstrate how well they work with our dataset and provide information on training setup. The image sequence and audio sequence inputs are denoted as $ x_{1:T}^{v} $ and $ x_{1:T}^{a} $ in all baselines, respectively.

\subsection{Automatic Speech Recognition (ASR)}
    The present section reviews the experiments conducted on speech recognition using the produced datasets. Audio data alone was used in the first experiment to train Persian speech recognition. To this end, the pre-trained AV-HuBERT~\cite{shi2022avhubert} model uses $ x_{1:T}^{a} $ to extract a 768-dimensional feature vector $ e_{1:T}^{a} $  for each 40ms of audio input. In this experiment, we used the pre-trained AV-HuBERT base model for feature extraction on the English language data. The output embedding in this network for each window is a representation of that audio segment, including speaker and speech features. We expect the model to use speech-related features during the learning process. As such, we expect features corresponding to the speaker to be implicitly ignored, as they are irrelevant to our purpose. One further consideration during the experiment is whether or not a model pre-trained via the English language data can be used for the Persian language.\\
    In the next step, the $ e_{1:T}^{a} $ vector was fed to a two-stacked Bidirectional LSTM, resulting in $ o_{1:T}^{a} $. This network was trained using CTC. We assume $ y = (y_1, y_2, ..., y_n) $ is a transcription and $ \pi $ are paths that function B will map $ \pi $ to y. We calculated the $ p_t^{CTC} $ probability over 52 characters by applying Softmax on $ o_{1:T}^{a} $. As our CTC loss $ L_{CTC} $ 
    % (According to Equation \ref{first_equ})
    decreases, the network begins to learn.\\
    As evident in table \ref{tab:results_avsr}, using the proposed architecture, we achieved a \%84.66 character accuracy rate (CAR) using only 211 hours of Persian speech data, which is a suitable value, considering the data volume. Moreover, the experiment has demonstrated that the AV-HuBERT model, trained on data in English, could extract a proper embedding for speech data in Persian.

    % \begin{equation}\label{first_equ}
    % L_{CTC} = -\log\sum_{\pi\in B^{-1}(y)}\prod_{t=1}^Tp_t^{CTC}(\pi_t|e_{1:T}^{a})
    % \end{equation}

\subsection{Audio Visual Speech Recognition (AVSR)}
\begin{figure*}[t]
  \centering
  \includegraphics[width=0.9\textwidth]{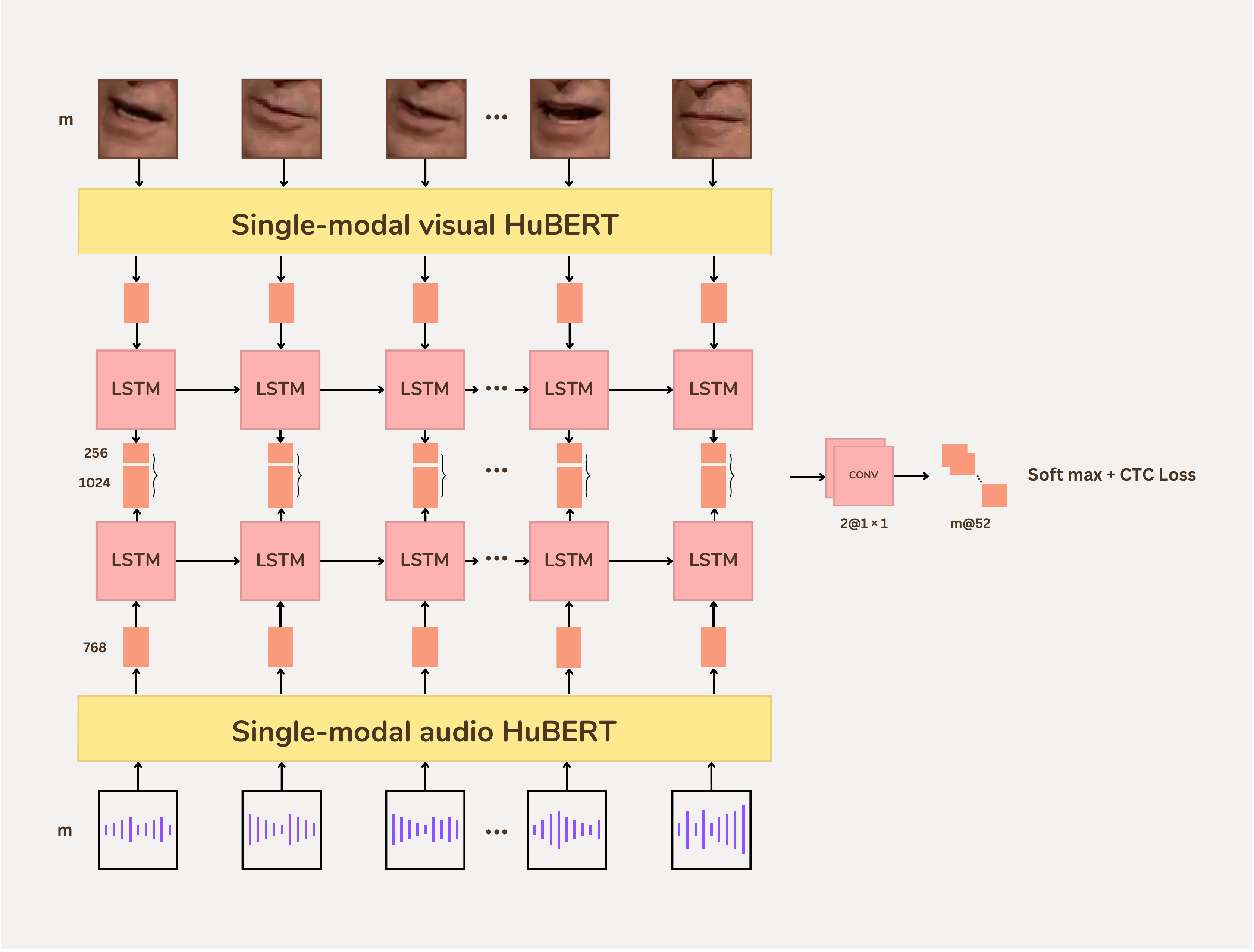}
  \caption{The audio-visual architecture.}
  \label{fig:arc_avsr}
\end{figure*}

    During the second experiment, we trained an audio-visual speech recognition model. Image pre-processing was first used in this model to crop only the mouth ROI. Then the AV-HuBERT model was used per cropped video frame to obtain a 768-dimensional vector. In this section, an architecture similar to the previous stage was used, but we also gave the visual features as input to the model. Moreover, we used less output for visual features as they presumably contain less information than speech information. This model is presented in Figure \ref{fig:arc_avsr}.  The input image sequence $ x_{1:T}^{v} $ is fed to the pre-trained Single-modal visual HuBERT to obtain a 768-dimensional feature vector $ e_{1:T}^v $. The 768-dimensional feature vector $ e_{1:T}^a $ is obtained from input audio sequence $ x_{1:T}^{a} $ through the pre-trained Singel-modal audio HuBERT concurrently. In the next step, the separated two-stacked Bi-LSTM are used to map $ e_{1:T}^v $ and $ e_{1:T}^a $ to $ o_{1:T}^v $ and $ o_{1:T}^a $ respectively. Then the two 1-dimensional convolutional filters are applied on $ c_t $ as a result of the concatenation of $ o_t^v $ and $ o_t^a $.
    % (Equation \ref{second_equ} and \ref{third_equ}). 
    The output of this step is $ f_t $. As mentioned above, CTC loss is used to train the model. First, we calculated the $ p_t^{CTC} $ probability. 
    % According to the equation \ref{4th_equ},
    This probability is obtained by applying Softmax and linear projection on $ f_t $ (Where $ W \in {\rm I\!R^{52\times (256 + 1024)}} $ and $ b \in {\rm I\!R^{52}} $). Then, the model will be optimized through $ L_{CTC} $ minimizing the value. 
    % (Equation \ref{5th_equ}).
    
    % \begin{equation}\label{second_equ}
    % c_t = Concat(o_t^v, o_t^a)
    % \end{equation}

    % \begin{equation}\label{third_equ}
    % f_t = Conv1D(c_t)
    % \end{equation}

    % \begin{equation}\label{4th_equ}
    % p_t^{CTC} = Softmax(Wf_t + b)
    % \end{equation}

    % \begin{equation}\label{5th_equ}
    % L_{CTC} = -\log\sum_{\pi\in B^{-1}(y)}\prod_{t=1}^Tp_t^{CTC}(\pi_t|f_{1:T})
    % \end{equation}
    
    This model was trained with the exact mechanism as the previous one.  As visible in table \ref{tab:results_avsr}, through this model, which provides both audio and visual features, we have achieved higher accuracy than the ASR model. As visible in this experiment, embeddings obtained in the pre-trained AV-HuBERT model, which are trained on the English language data, can be used in the Farsi language, and relatively desirable features of lip movement can be extracted from it.

\begin{table*}[ht]
 \caption{Speech recognition results on our dataset }
%  \vspace{2mm}
  \centering
  \begin{adjustbox}{width=1\textwidth}
  \begin{tabular}{c | c |c |c|c}
    \toprule
    \toprule
    Method  & CER (Greedy decoding)  & WER (Greedy decoding) & CER (Beam search decoding)  & WER (Beam search decoding) \\
    \midrule
     ASR(CTC) & \%15.77 & \%46.43  & \%15.34  & \%45.86   \\
     AVSR(CTC) & \textbf{\%13.76} &  \textbf{\%44.98} & \textbf{\%13.08} & \textbf{\%43.43}   \\
    \bottomrule
    \bottomrule
  \end{tabular}
  \label{tab:results_avsr}
  \end{adjustbox}
\end{table*}

\subsection{Speaker recognition}
Another problem we tested using the generated data was speaker recognition. For this, we first separated the people with more than five samples. Then we randomly selected five samples for each speaker. For each speaker, four samples were considered as a reference, and only one of the samples was used for testing.
Then, using the TitaNet~\cite{koluguri2022titanet} pre-trained model, which was trained on 5 datasets (Voxceleb~\cite{Nagrani17} and Voxceleb2~\cite{Chung18b}, NIST SRE portion of datasets
from 2004-2008 (LDC2009E100), Switchboard-Cellular1
and Switchboard-Cellular2~\cite{godfrey1993switchboard}, Fisher~\cite{cieri2004fisher} and Librispeech~\cite{panayotov2015librispeech}).
We fed $ x_{1:T}^{a_{t}} $ as a test and $ x_{1:T}^{a_{0}} $, ..., $ x_{1:T}^{a_{3}} $ as references to the TitaNet~\cite{koluguri2022titanet} model. The model produces the 192-dimensional embeddings of each audio sample, for example, $ v_{1:T}^{a_{t}} $. In the next step, the test embedding vector is compared with every reference of every person through cosine similarity
% according to the equation \ref{6th_equ} 
(Here $ i \in {0, ..., 3}$). After that, we calculate the average with the previous step's results.
% (Equation \ref{7th_equ}). 
Then, each sample in the test was assigned to the person who had the most similarity with that in terms of the average cosine similarity criterion. Using this method, we achieved an accuracy of 84.5.

% \begin{equation}\label{6th_equ}
% S_{c_{i}}(v^{a_{t}}, v^{a_{i}}) = \frac{v^{a_{t}}.v^{a_{i}}}{||v^{a_{t}}||||v^{a_{i}}||}
% \end{equation}

% \begin{equation}\label{7th_equ}
% m = \frac{\sum_{i=0}^{3}S_{c_{i}}}{4}
% \end{equation}

\section{Conclusion}
Arman-AV is a large-scale multipurpose dataset for Persian that can be used in various tasks such as lip reading, automatic speech recognition, audio-visual speech recognition, and speaker recognition, and it is publicly available. The dataset consists of almost 220 hours of video data (about 89,000 samples) from 1760 speakers. Although the main goal of collecting this dataset was lip reading, the results of a baseline method were reported for each of the mentioned tasks. According to the obtained results, the character error rate of speech recognition was reduced by 14.7\% relatively using visual features. We also proposed a method for obtaining Persian visemes. By using these visemes, higher accuracy in lip reading was achieved. In addition, we have provided various analyses for the dataset such as the age and gender of speakers, and the synchronization of audio and video for each segment. All of this metadata is available along with the dataset. The dataset can be used for other tasks such as face recognition, face verification, and audio-visual speaker recognition, which we did not cover in this paper and can be considered for future work on the dataset.

\section{Acknowledgement}

The project was mainly supported by Arman Rayan Sharif company, an AI company in Iran.

\bibliographystyle{IEEEtran}

\bibliography{mybib}

\end{document}